\def\year{2021}\relax
\documentclass[letterpaper]{article} 
\usepackage{aaai21}  
\usepackage{times}  
\usepackage{helvet} 
\usepackage{courier}  
\usepackage[hyphens]{url}  
\usepackage{graphicx} 
\usepackage{natbib}  
\usepackage{caption} 
\usepackage{booktabs}
\usepackage{tabularx}
\usepackage{multirow}
\usepackage{amssymb}
\usepackage{amsmath}
\usepackage{enumitem}
\usepackage{latexsym}
\usepackage{eucal}
\usepackage{bm}
\usepackage{amsfonts}
\usepackage{subfigure}
\usepackage{enumerate}

\usepackage[switch]{lineno}

\title{Generalized Relation Learning with Semantic Correlation Awareness\\ 
for Link Prediction}
\author {
        Yao Zhang\textsuperscript{\rm 1},
        Xu Zhang\textsuperscript{\rm 1},
        Jun Wang\textsuperscript{\rm 2},
        Hongru Liang\textsuperscript{\rm 1},\\
        Wenqiang Lei\textsuperscript{\rm 3}\footnote{Corresponding authors.},
        Zhe Sun\textsuperscript{\rm 4},
        Adam Jatowt\textsuperscript{\rm 5},
        Zhenglu Yang\textsuperscript{\rm 1}\footnotemark[1]
        \\
}
\affiliations {
    \textsuperscript{\rm 1} TKLNDST, CS, Nankai University, China\\
    \textsuperscript{\rm 2} Ludong University, China
    \textsuperscript{\rm 3} National University of Singapore, Singapore\\
    \textsuperscript{\rm 4} Computational Engineering Applications Unit, RIKEN, Japan
    \textsuperscript{\rm 5} University of Innsbruck, Austria \\
    $\left \{\right.$yaozhang, xuzhang, lianghr, junwang$\left. \right \}$@mail.nankai.edu.cn, 
    wenqianglei@gmail.com,\\
    zhe.sun.vk@riken.jp,
    adam.jatowt@uibk.ac.at, 
    yangzl@nankai.edu.cn

}
\begin{document}
\maketitle

\begin{abstract}
 Developing link prediction models to automatically complete knowledge graphs has recently been the focus of significant research interest. The current methods for the link prediction task have two natural problems: 1) the relation distributions in KGs are usually unbalanced, and 2) there are many unseen relations that occur in practical situations. These two problems limit the training effectiveness and practical applications of the existing link prediction models. We advocate a holistic understanding of KGs and we propose in this work a unified Generalized Relation Learning framework GRL to address the above two problems, which can be plugged into existing link prediction models. GRL conducts a generalized relation learning, which is aware of semantic correlations between relations that serve as a bridge to connect semantically similar relations. After training with GRL, the closeness of semantically similar relations in vector space and the discrimination of dissimilar relations are improved. We perform comprehensive experiments on six benchmarks to demonstrate the superior capability of GRL in the link prediction task. In particular, GRL is found to enhance the existing link prediction models making them insensitive to unbalanced relation distributions and capable of learning unseen relations. 
\end{abstract}



\section{Introduction}
\label{intro}

\begin{figure}[t]

\centering
\includegraphics [width=7cm]
{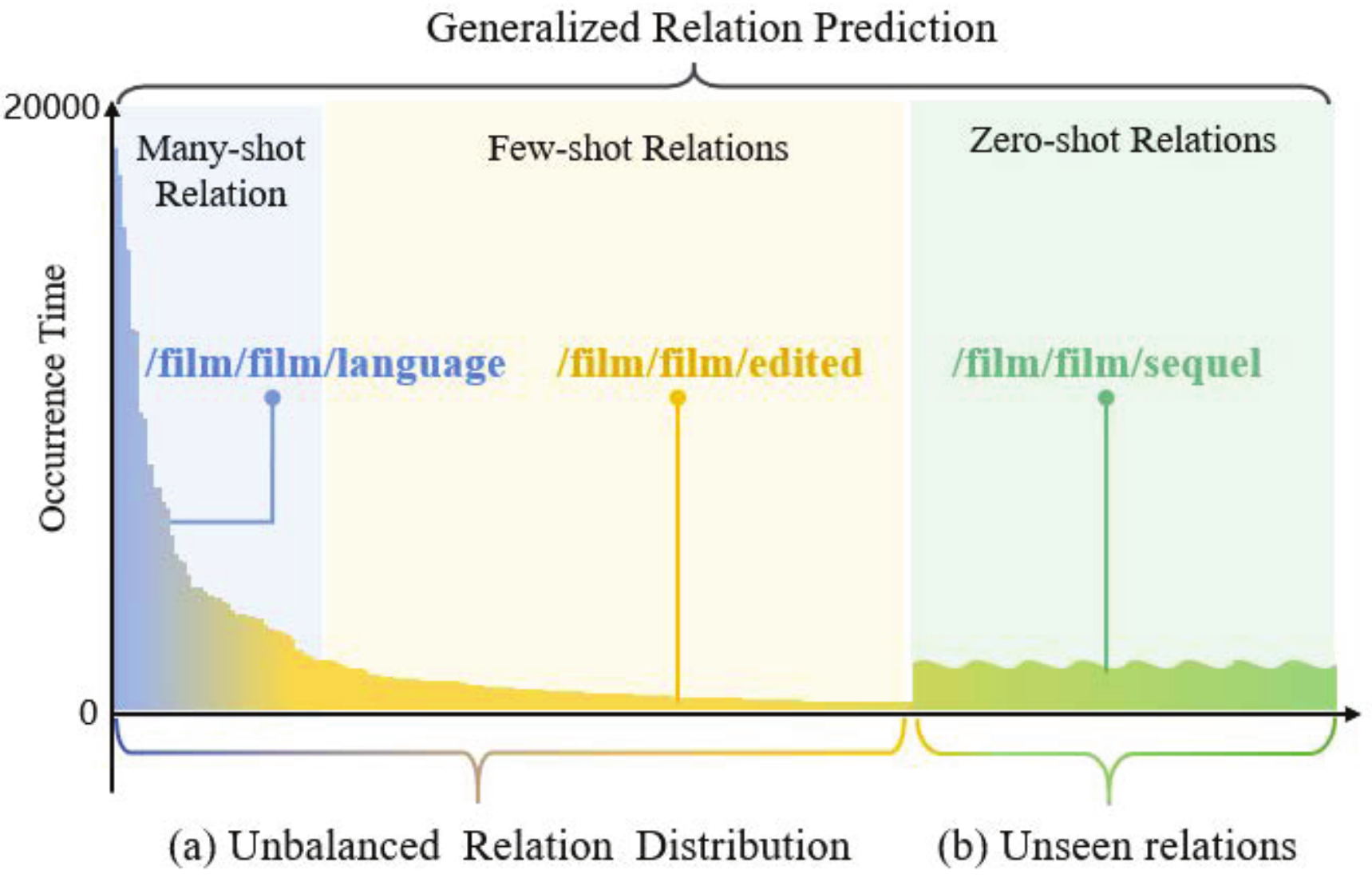}
\caption{ (a) The unbalanced relation distribution in the FB15K-237 dataset where relations are sorted according to their frequency. (b) Lots of unseen relations. Three \textit{film}-related relations are respectively 
categorized into the many-shot class, few-shot class, and zero-shot class as marked. 
}

\label{fig:intro}      
\end{figure}

Knowledge graphs~(KGs), representing facts in semantic graph structures, have been applied
to multiple artificial intelligence tasks, e.g., recommendation~\cite{leiconversational,2021Learning}, dialogue generation~\cite{moon-etal-2019-opendialkg,leiinteractive}, and question answering~\cite{ChristmannRASW19,zhu2021retrieving}.
In KGs, facts are formed as triples, \texttt{(head~entity, relation, tail~entity)}, where the head entity is linked to the tail entity via the relation.
New knowledge 
emerges continuously,
and hence the issue of incompleteness of KGs has triggered 
wide research interests 
in 
link prediction task, 
which requires predicting the missing links in KGs~\cite{simple}. 
The mainstream link prediction models~\cite{bordes2013translating,Dettmers2018Convolutional} 
learn the embeddings of entities and relations, and then use a score function to estimate the validity of triples.
However, we 
believe using the embedding learning for mainstream link prediction models results in two key problems:
\begin{itemize}[leftmargin=*]
    \item[\small$\bullet$] \textit{Unbalanced relation distribution. 
    } 
    As shown in Figure~\ref{fig:intro}, 
    the relation distribution in an off-the-shelf KG learning resource~(i.e., FB15K-237~\cite{toutanova-chen-2015-observed}) is quite unbalanced.
    For example, {
    the frequencies of the two relations \texttt{/film/film/language}
    and \texttt{/film/film/edited}
    differ greatly.}
    Mainstream link prediction models assume enough training instances for all relations and 
    pay less attention to few-shot relations, 
    disregarding the fact that few-shot relation learning may 
    influence the learning performance to a high degree.

    \item[\small$\bullet$] \textit{Existence of unseen relations.
    } 
    Real-world KGs tend to be open and evolve quickly,
    and accordingly there is a large number of zero-shot relations unseen in the off-the-shelf learning resources, for example, the relation {\texttt{/film/film/sequel}} in Figure~\ref{fig:intro}.
    The unseen relations are beyond the capacity of mainstream link prediction models, as there are no training instances to learn their embeddings.
    This problem 
    may restrict the use of these models in 
    downstream tasks.
\end{itemize}

Recently, some efforts have been conducted on addressing the above problems.
\citet{Xiongoneshot}, \citet{shi2018open}, and \citet{ChenMeta} 
adopted the meta-learning or metric-based approaches to train 
on 
limited training samples and perform fast learning on new few-shot relations. 
These studies 
show promise in few-shot relation learning, 
however they have difficulty in tackling
unbalanced relation distributions, 
which is mainly attributed to the excessive time cost required for training numerous relations.
More recently, 
\citet{DBLP:journals/corr/abs-1907-08937}, \citet{Qin2020GenerativeAZ} predicted 
the unseen relations 
by extracting information from textual descriptions.
They were able to successfully complete 
the unseen relation prediction task. However, these models are not appropriate for 
link prediction task,
since textual descriptions tend to be noisy and 
also cannot build a bridge 
between seen and unseen relations.
In general, an ideal link prediction model should be able to jointly learn many-, few-, and zero- shot relations.



Regarding the joint relation learning, we noticed that semantic correlations, which denote the similarities of relations
in semantics, can 
serve as a bridge to connect 
the learning of many-, few-, and zero- shot relations.
Take Figure~\ref{fig:intro} as an instance.
The many-shot relation ``\texttt{/film/film/language}'', few-shot relation  ``\texttt{/film/film/edited}'', and zero-shot relation ``\texttt{/film/film/sequel}'' are all related to ``\texttt{film}''.
Based on the assumption that semantically similar relations should be located near each other 
in embedding space~\cite{yang2014embedding}, it makes sense 
to exploit semantic correlations, such as the one in the above-mentioned example, to 
accomplish the joint relation learning.
Inspired by 
this, we propose 
a \textbf{G}eneralized \textbf{R}elation \textbf{L}earning framework (abbreviated to GRL)
based on learning semantic correlations.
GRL can be plugged into a mainstream link prediction model to make it (1) insensitive to unbalanced relation distributions and (2) capable of learning zero-shot relations.

Specifically, GRL is plugged into a link prediction model
after the embedding learning stage. 
To optimize the relation embedding, GRL extracts rich semantic correlations through an attention mechanism, fuses different relations, and minimizes the classification-aware loss to enable these 
implicitly embedded semantic correlations in the relation embeddings. 
Then, the closeness of semantically similar relations in vector space and the discrimination of dissimilar relations can be improved.
In this way, few-shot relations can learn knowledge from the semantically similar many-shot relations;
for zero-shot relations, their most semantically similar relation can also be predicted.
In our experiments, we improve two base models (DistMult~\cite{yang2014embedding} and ConvE~\cite{Dettmers2018Convolutional}) 
by incorporating the proposed GRL framework on all relation classes, i.e., many, few, and zero-shot relations.
Our work is an important step towards a holistic understanding of KGs and a generalized solution of relation learning for the link prediction task.

Our contributions are as follows:

    \begin{itemize}[leftmargin=*]
        \item[\small$\bullet$] We carefully consider two key problems of the embedding learning used by mainstream link prediction models and we highlight the necessity of jointly learning many-, few-, and zero- shot relations.

        \item[\small$\bullet$] We propose GRL framework by leveraging the rich semantic correlations between relations to 
        make the link prediction models insensitive to unbalanced relation distributions and capable of learning zero-shot relations.

        \item[\small$\bullet$] We perform experiments on six benchmarks to 
        evaluate the link prediction capability of GRL,
        and show that GRL lets the base link prediction models perform well across many-, few-, and zero- shot relations.

        \end{itemize}

        \section{Related Work}
        \label{relatedwork}
        
        Since KGs are populated by automatic text processing they are often incomplete, and it is usually infeasible to manually add to them all the relevant facts. Hence, many researches approached the task of 
        predicting missing links in KGs.
        
        \textbf{Mainstream link prediction models} widely use embedding-based methods to map entities and relations into continuous low-dimensional vector space and use a score function to predict whether the triples are valid. They can be broadly classified as translational based~\cite{bordes2013translating,wang2014knowledge, lin2015learning, ji2016knowledge}, multiplicative based~\cite{nickel2011three,yang2014embedding,trouillon2017complex}, and neural network based~\cite{Dettmers2018Convolutional,schlichtkrull2018modeling}.
        These models are based on the implicit assumption that all relations are distributed within the dataset in a balanced way.
        Hence, they perform poorly
        in few-shot relation learning scenarios because these models neglect the imbalanced distributions, as well as they cannot properly handle 
        zero-shot relations
        due to keeping only the knowledge of existing relations and not learning information on unseen relations.

        \textbf{Few-shot relation learning models} attempt to adopt the meta-learning~\cite{ChenMeta,lv2019adapting} and  metric-based~\cite{Xiongoneshot,wang2019tackling} methods to learn knowledge from only a few samples.
        However, the few-shot learning models are computationally expensive because 
        they need to spend extra time retraining on each few-shot relation (meta-learning), or need to compare the few-shot relations one by one (metric-based).
         In practice, the many-shot and few-shot scenarios are not explicitly distinguished.
        \textbf{Zero-shot relation learning models} aim to learn relations that are unseen in the training set. 
        Researchers have proposed several models to deal with zero-shot relations by leveraging information from textual descriptions \cite{DBLP:journals/corr/abs-1907-08937,Qin2020GenerativeAZ}. They perform well on predicting the zero-shot relations,
        but are not appropriate in the link prediction task because textual descriptions could be noisy and a bridge connecting seen and unseen relations could be missing.
        
        \begin{figure*}[t]

        \centering
        \includegraphics [width=13.9cm]
        {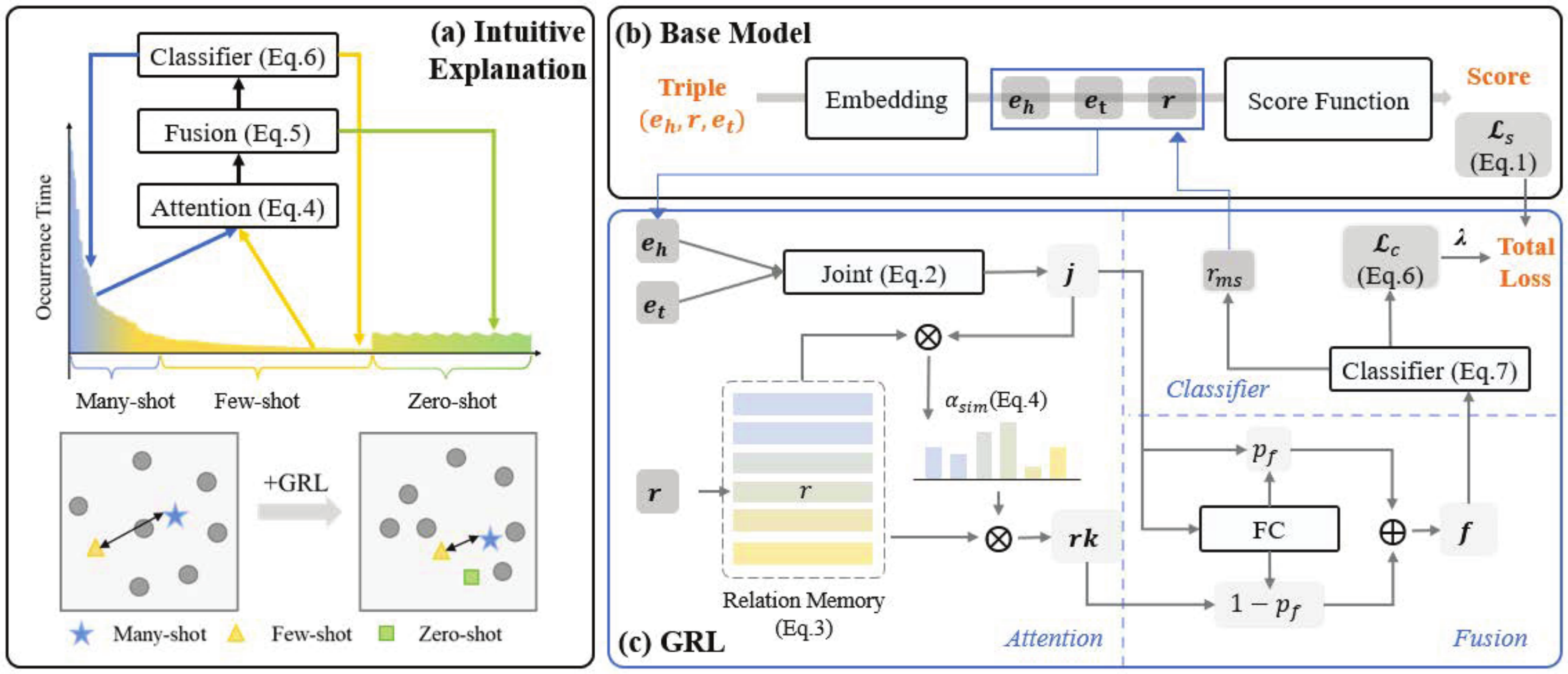}
        \caption{The illustration of GRL, which consists of the intuitive explanation \textbf{(a)}, the base model \textbf{(b)} and the detailed architecture \textbf{(c)}.
        The base model denotes the mainstream link prediction model.
        GRL is plugged after the embedding component of the base model and contains three components: attention, fusion, and classifier.
        }
        
        \label{fig:modeloverview}      
        \end{figure*}
        
        In this work, we 
        focus on jointly learning many-, few-, and zero- shot relations without requiring extra textual knowledge.
        Recently, some computer vision works~\cite{ye2019learning,RelationalGeneralized2019} 
        have 
        attempted to 
        approach the generalized image classification. 
        Nonetheless, they are 
        not designed for coping with
        graph structures, e.g., KGs.
        We leverage in this work the rich semantic correlations between relations as a bridge to connect the
        learning of many-, few-, and zero- shot relations.
        \citet{zhang2019long} integrated the rich semantic correlations between specific hierarchical relations into relation extraction. 
        That method performs well only on hierarchical relations, as well as it predicts relations from text, hence it does not cope with the link prediction task.
        

\section{Method}
\label{method}

Figure~\ref{fig:modeloverview} provides the illustration of the proposed framework GRL. The figure consists of three parts: the intuitive explanation of GRL~in Figure~\ref{fig:modeloverview} (a), base model shown in Figure~\ref{fig:modeloverview} (b), and  the detailed architecture
~in Figure~\ref{fig:modeloverview} (c).

The intuitive explanation of GRL is shown to utilize the semantic correlations between many-shot and few-shot relations so that the relation embedding learning can benefit from semantically similar relations.
We devise three modules, i.e., \textit{Attention}, \textit{Fusion} and \textit{Classifier}, to embed and fuse the rich semantic correlations among many-shot and few-shot relations in the training phase;
and to select the most similar relation embedding for zero-shot relations in the testing phase.
In this way, GRL can improve the performance on all relation classes, i.e., many, few, and zero-shot relations.
The base model denotes the existing mainstream link prediction model consisting of an embedding component and a score function component. 
GRL can be plugged between the embedding and the score function components to make it (1) insensitive to imbalanced relation distributions and (2) capable of detecting zero-shot relations.

Before 
delving into the model description, we first formally represent a KG as a collection of triples~$\mathcal{T}=\left \{ (e_{h},r,e_{t})|e_{h}\in \mathcal{E}, e_{t}\in \mathcal{E},r\in \mathcal{R}\right \} $, where $\mathcal{E}$ and $\mathcal{R}$ are the entity and relation sets, respectively. Each directed link in KG represents a triple~(i.e., $e_{h}$ and $e_{t}$ are represented as the nodes and $r$ as the labeled edge between them).
The link prediction task is 
to predict whether a given triple~$(e_{h},r,e_{t})$ is valid or not.
In particular, for the zero-shot relations, we need to emphasize that we mainly focus on predicting the validity of the triple with a zero-shot relation, rather than predicting the zero-shot relations, i.e., the relation prediction task~\cite{DBLP:journals/corr/abs-1907-08937,Qin2020GenerativeAZ}.
However, GRL has also the ability to predict the most semantically similar relation of a given zero-shot relation through learning from the many- and few-shot relations, not from the text description.



\subsection{Base Model}
\label{baselingmodel}


We select a mainstream link prediction model as the base model and apply GRL to it.
The base model can be seen as multi-layer neural network consisting of an embedding component and a score function component.
For the base link prediction model, given an input triple~$(e_{h},r,e_{t})$, the embedding component maps the head and tail entities~$(e_{h},e_{t})$ and the relation~$r$ to their distributed embedding representations~$(\bm{e}_{h},\bm{r},\bm{e}_{t})$ through the entity and relation embedding layers, respectively. After the embedding representations are obtained, the score function component is adopted to calculate the likelihood of $(\bm{e}_{h},\bm{r},\bm{e}_{t})$ being a valid fact. The following binary cross entropy loss is used to train model parameters:
\begin{equation} 
    \mathcal{L}_{s}=-\frac{1}{N}\sum_{i=1}^{N}(t_{i}\log  p(s_{i})+(1-t_{i})\log(1-  p(s_{i}))),
\end{equation}
where $s_{i}$ is the score of the $i$-th input triple, $t_{i}$ is the ground truth label, $t_{i}$ is 1 if the input relation is valid and 0 otherwise, and $N$ is the number of input triples.


\subsection{GRL Framework}
\label{sec:mffl}
The loss used by mainstream link prediction models is score-oriented and lacks an in-depth exploration of rich semantic correlations in KGs. We propose the GRL framework to learn appropriate representations for relations by embedding semantic correlations into classification-aware optimization. 
GRL contains three specific modules:

1) \textit{Attention} Module, which builds the knowledge-aware attention distribution and the relational knowledge vector. The aim of this module is to extract the semantic correlations and the degree of these correlations.

2) \textit{Fusion} Module, which fuses the relational knowledge vector with the joint vector obtained from the attention module. This module realizes the fusion of different relations, according to semantic correlations.

3) \textit{Classifier} Module, which calculates the classification-aware loss to implicitly enable the rich semantic correlations embedded in the embeddings. Thanks to it, both the compactness of semantically similar relations and discrimination of dissimilar relations can be enhanced.

The following is a detailed introduction to each module.

\noindent
\textbf{Attention Module}

\textit{Joint Block.}
The classification-aware loss is calculated by the relation classification results based on the head and tail entities from the given triple~$(e_{h},r,e_{t})$.
Inspired by~\cite{Qin2020GenerativeAZ}, the joint vector of the head and tail entities has the ability to represent the potential relation between them.
The head and tail entities representations~(i.e., $\bm{e}_{h}$ and $\bm{e}_{t}$) are jointed together at the joint block for which we adopt three different alternatives:
\begin{equation}
    \bm{j}= 
\left\{
    \begin{array}{ll}
        \bm{e}_{h}-\bm{e}_{t}, & sub\\
        \bm{e}_{h}\otimes \bm{e}_{t}, & multiply\\
        W_{1}[\bm{e}_{h};\bm{e}_{t}]+b_{1}, & concat 
    \end{array},
\right.
\label{eq:joint}
\end{equation}
where $\otimes$ denotes the element-wise multiplication operator, and ~$W_{j}$ and $b_{j}$ are the learnable parameters.

\textit{Relation Memory Block.} Using a memory block to store 
class information is widely used in image classification 
\cite{SnellSZ17,RepMet2019,liu2019large}.
Following these 
studies, we design a relation memory block to store all relation information by sharing parameters with the relation embedding layer as
\begin{equation} 
    \bm{M}=\left \{\bm{r}_{1},\bm{r}_{2},...,\bm{r}_{K-1},\bm{r}_{K}\right \},
   \end{equation}
where $M\in \mathbb{R}^{K \times dim}$, 
$K$ is the number of relation classes.
As the training progresses, the relation embedding layer and relation memory block are updated synchronously.

\textit{Relational Knowledge.}
To realize the classification-aware optimization objective, we extract useful relational knowledge from the relation memory block to enrich the joint vector.
The semantic correlation degree between different relations may vary; thus, we adopt the attention mechanism to customize specific relational knowledge for each joint vector.
Concretely, the relational knowledge vector~$\bm{rk}$ is computed as a weighted sum of each relation representation in the relation memory block~$M$, i.e., $\bm{rk}=\alpha _{sim}\bm{M}$, 
where $\alpha _{sim}\in \mathbb{R}^{K}$ represents the knowledge-aware attention distribution.

\textit{Attention Distribution}
The knowledge-aware attention distribution $\alpha _{sim}$ describes the similarity between the  joint vector and  each relation representation in the relation memory block.
We 
estimate $\alpha _{sim}$ as 
\begin{equation} 
    \alpha _{sim}=softmax(\bm{j}\bm{M}^{\top}),  
    \label{eq:sim}
\end{equation}
where
$softmax$ is the activation function, and $\bm{M}^{\top}$ represents the transposed matrix of $\bm{M}$.
Note that the attention value of the ground-truth relation is masked with 0.

\noindent
\textbf{Fusion Module}

In this module, the joint vector and relational knowledge vector are fused.
Intuitively, the proportion of fusion is different for each joint vector.
Inspired by the pointer-generator network~\cite{pointergenerator} 
that facilitates 
copying words from the source text during 
new words 
generation, we propose a soft switch, that is, the fusion probability~$p_{f}\in [0,1]$, to adaptively adjust the fusion proportion between the joint vector and relational knowledge vector.
The fusion probability~$p_{f}$ is 
estimated according to the joint vector as $p_{f}=sigmoid(FC(\bm{j})),$
where~$FC$ is the fully connected neural network, and $sigmoid$ is the activation function.
Finally, we obtain the following fusion vector~$\bm{f}$ over the joint vector~$\bm{j}$ and relational knowledge vector~$\bm{rk}$ as
\begin{equation} 
    \bm{f}=(1-p_{f})\bm{j}+p_{f}\bm{rk}.
   \end{equation}

\noindent
\textbf{Classifier Module}

\textit{Classification-aware Loss.}
The fusion vector~$\bm{f}$ 
is mapped to a class 
probability through the classifier block as
\begin{equation} 
    D \sim  softmax(\bm{f}^{\top}W_{c} ),
   \end{equation}
where $W_{c}\in \mathbb{R}^{dim \times K}$ is the classification weight matrix, and $softmax$ is the activation function. 

Given the ground truth relation~$r_{i}$ from the $i$-th input~$(e_{h_{i}},r_{i},e_{t_{i}})$, we adopt 
cross entropy to 
assess the classification-aware loss as
\begin{equation} 
    \mathcal{L}_{c}=-\frac{1}{N}\sum_{i=1}^{N}\log  p(r_{i}|(e_{h_{i}},e_{t_{i}})),
\end{equation}
where $p(r_{i}|(e_{h_{i}},e_{t_{i}}))\in D_{i}$ is the probability of the ground truth relation~$r_{i}$.




\begin{table}

    \begin{center}
    
    \small
    \begin{tabular}{l|ccccc}
    
    \toprule[1.2pt]
    &    $\mathcal{\left|E\right|}$   &     $\mathcal{\left|R\right|}$    &    Train   &     Valid   &     Test\\
    \midrule[0.6pt]
    YAGO3-10   &     123k    &   37  &     1M   &     5k   &   5k\\
    FB15K-237   &     15k    &   237  &     273k   &     18k   &   20k\\
    NELL-995   &     75k    &   200  &     150k   &     543   &   4k\\
    Kinship   &     104    &   25  &     9k   &     1k   &   1k\\
    WN18   &     41k    &   18  &     141k   &     5k   &   5k\\
    NELL-ONE   &     69k    &   358  &     190k   &     1k   &   2k\\
    \bottomrule[1.2pt]
    
    \end{tabular}
    \end{center}
    \caption{Statistics of datasets. $\mathcal{\left|E\right|}$ and $\mathcal{\left|R\right|}$ represent the cardinalities of the  entity and relation sets.}
    \label{table:datasets} 

    \end{table}

\begin{table*}              
    \begin{center}
    
    \small
    \begin{tabular}{lm{0.58cm}<{\centering}m{0.58cm}<{\centering}m{0.75cm}<{\centering}|m{0.58cm}<{\centering}m{0.58cm}<{\centering}m{0.75cm}<{\centering}|m{0.58cm}<{\centering}m{0.58cm}<{\centering}m{0.75cm}<{\centering}|m{0.58cm}<{\centering}m{0.58cm}<{\centering}m{0.75cm}<{\centering}|m{0.58cm}<{\centering}m{0.58cm}<{\centering}m{0.75cm}<{\centering}}

    \toprule[1.2pt]
    \multirow{3}{*}{}  & \multicolumn{3}{c|}{YAGO3-10}  & 
    \multicolumn{3}{c|}{FB15K-237}    & \multicolumn{3}{c}{NELL-995}& \multicolumn{3}{c|}{Kinship} & \multicolumn{3}{c}{WN18} \\ \cline{2-16}\rule{0pt}{12pt}
    & \multirow{2}{*}{MRR} &  \multicolumn{2}{c|}{HITS@N} & \multirow{2}{*}{MRR} & \multicolumn{2}{c|}{HITS@N} & \multirow{2}{*}{MRR} &  \multicolumn{2}{c|}{HITS@N} & \multirow{2}{*}{MRR} &  \multicolumn{2}{c|}{HITS@N} & \multirow{2}{*}{MRR} & \multicolumn{2}{c}{HITS@N} 
    \\ \cline{3-4}\cline{6-7}\cline{9-10}\cline{12-13}\cline{15-16}\rule{0pt}{10pt}
    & & @10  &   @1  & & @10  &   @1  & & @10  &   @1  & & @10  &   @1  & & @10  &   @1  \\
    \midrule[1.2pt]
    ComplEx
    &36.0&55.0&26.0
    &24.7&42.8&15.8
    &48.2&60.6&39.9
    &82.3&97.1&73.3
    &94.1&94.7&93.6
    \\
    R-GCN
    &-&-&-
    &24.8&41.7&15.3
    &12.0&18.8&8.2
    &10.9&23.9&3.0
    &81.4&\textbf{96.4}&69.7
    \\
    ConvKB  
    &-&-&-
    &28.9&47.1&19.8
    &43.0&54.5&37.0
    &61.4&95.3&43.6
    &-&-&-
    \\
    D4-STE
    &47.2&64.3&38.1
    &32.0&\textbf{50.2}&23.0
    &-&-&-
    &-&-&-
    &94.6&95.2&94.2
    \\
    D4-Gumbel
    &38.8&57.3&29.4
    &30.0&49.6&20.4
    &-&-&-
    &-&-&-
    &94.6&95.4&94.2
    \\\midrule[1.2pt]
    DistMult
    &34.0&54.0&24.0
    &24.1&41.9&15.5
    &48.5&61.0&40.1
    &51.6&86.7&36.7
    &82.2&93.6&72.8
    \\
    \multirow{1}{*}{+GRL}
    &41.2&59.9&31.1
    &25.8&43.9&16.9
    &\textbf{54.3}&\textbf{64.6}&\textbf{47.6}
    &52.2&86.4&37.3
    &86.1&95.2&79.2
    \\
    ($\pm$ sd)&{(0.3)}&{(1.0)}&{(0.1)}
    &{(0.2)}&{(0.3)}&{(0.1)}
    &{(0.2)}&{(0.3)}&{(0.3)}
    &{(0.2)}&{(0.8)}&{(0.2)}
    &{(1.0)}&{(0.4)}&{(1.1)}
    \\
    \midrule[0.6pt]
    ConvE
    &52.0&66.0&45.0
    &31.6&49.1&23.9
    &49.1&{61.3}&40.3
    &83.3&\textbf{98.1}&73.8
    &94.2&95.5&93.5
    \\
    \multirow{1}{*}{+GRL}
    &\textbf{55.4}&\textbf{69.0}&\textbf{47.4}
    &\textbf{32.6}&\textbf{50.2}&\textbf{24.8}
    &{49.4}&60.6&{41.5}
    &\textbf{83.4}&97.8&\textbf{74.5}
    &\textbf{94.8}&{95.7}&\textbf{94.4}
    \\
    ($\pm$ sd)&{(1.0)}&{(0.1)}&{(0.1)}
    &{(0.3)}&{(0.2)}&{(0.3)}
    &{(0.2)}&{(0.3)}&{(0.3)}
    &{(0.2)}&{(0.5)}&{(0.5)}
    &{(0.1)}&{(0.4)}&{(0.0)}
    \\

    \bottomrule[1.2pt]
    \end{tabular}
    
    \end{center}
    \caption{
    Link prediction 
    results (mean $\pm$ sd) 
    of the compared models (\%): the best results are marked in bold~(pairwise t-test at 5\% significance level).
    }
    \label{table:mainresult}

    \end{table*}

    \begin{table}[t]
      \begin{center}
      
          \small
          
              \begin{tabular}{lm{0.58cm}<{\centering}m{0.58cm}<{\centering}m{0.58cm}<{\centering}|m{0.58cm}<{\centering}m{0.58cm}<{\centering}m{0.58cm}<{\centering}}
              
              \toprule[1.2pt]
              \multirow{2}{*}{} & \multicolumn{3}{c|}{YAGO3-10} & \multicolumn{3}{c}{NELL-995} \\ \cline{2-7}\rule{0pt}{12pt}
              & {Many}  & {Few}& {All}& {Many}  & {Few}& {All}\\ 
              
              \midrule[1.2pt]
              DistMult		&	38.1	&	26.7	&	34.0	&	52.6    &	41.9	&	48.5\\
              \multirow{1}{*}{DistMult+GRL}&	44.8	&	34.2	&	41.2	&	57.3   &	48.8	&	54.3\\
                  (Increment)&   ($\uparrow$6.7)&   ($\uparrow$7.5)&   ($\uparrow$7.2)&   ($\uparrow$4.7)&    ($\uparrow$6.9)&   ($\uparrow$5.8)\\
          \midrule[0.6pt]
      
              ConvE	&	57.9	&	20.0	&	52.4	   &	52.0	&	42.2&	49.1 \\
              \multirow{1}{*}{ConvE+GRL}   &	59.4	&	24.6	&	55.4	   &	52.4	&	43.9&	49.4 \\
              (Increment) &   ($\uparrow$1.5)&   ($\uparrow$4.6)&   ($\uparrow$3.0)&    ($\uparrow$0.4)&   ($\uparrow$1.7)&   ($\uparrow$0.3)\\
              \bottomrule[1.2pt]
              
              \end{tabular}
              \caption{Link prediction results with the increment~(\%) on many-shot and few-shot sub-groups, and entire test set.}
      \label{table:headtail} 
      \end{center}

      \end{table}

\textit{Most Similar Relation.}
Existing mainstream link prediction models have achieved impressive performance, 
yet they can only learn the patterns observed in the closed datasets, 
thereby 
limiting their scalability for handling the rapidly evolving KGs. 
Specifically, when a zero-shot relation~$r_{z}$~(i.e., one not existing in the training set) occurs between an entity pair~$(e_{h},e_{t})$, it is almost impossible for the existing models to distinguish whether 
this new 
triple~$(e_{h},r_{u},e_{t})$ 
is valid or not.
All $r_{z}$ will be then identified as an `unknown` vector~$\bm{u}$ by the embedding component,
and 
the 
newly constructed triple representation~$(\bm{e}_{h},\bm{u},\bm{e}_{t})$ will 
receive a low score.
To alleviate this defect, GRL selects the most semantically similar relation for replacing to enhance the learning ability of base model on zero-shot relations. 
We argue that the relation 
which corresponds to the maximum similarity in $\alpha _{sim}$
reflects the semantic relation of two entities in the best way.
Therefore, we use the vector of the most similar relation~$\bm{r}_{ms}$ to replace the vector $\bm{u}$ and 
evaluate the newly constructed triple representation~$(\bm{e}_{h},\bm{r}_{ms},\bm{e}_{t})$.

\subsection{Learning Scheme}
\label{learningscheme}
We follow the definition of score-aware loss 
in existing base models and 
propose a classification-aware loss to train the model.
The overall optimization follows the joint learning paradigm that is defined as a weighted combination of constituent losses as $\mathcal{L}=\mathcal{L}_{s}+\lambda \mathcal{L}_{c},$
where $\lambda$ is a hyper-parameter to 
balance the importance between the score-aware loss and classification-aware loss for optimization.


    \section{Experiments and Results}
    \label{experiments}
    \subsection{Datasets}
    \label{datasets}
    
    We select two categories of datasets to comprehensively evaluate GRL as follows, whose statistical descriptions are shown in Table~\ref{table:datasets}:
    
    \begin{itemize}[leftmargin=*]
        \item[\small$\bullet$] \textit{Imbalanced datasets}: YAGO3-10~\cite{Mahdisoltani2013YAGO3}, FB15K-237~\cite{toutanova-chen-2015-observed}, NELL-995~\cite{Xiong2018DeepPath}, Kinship~\cite{LinEMNLP2018}, and WN18~\cite{bordes2013translating}. These datasets contain both many-shot and few-shot relations.
        
        \item[\small$\bullet$] \textit{Few-shot dataset}: NELL-ONE~\cite{Xiongoneshot}, which is specially constructed for the few-shot learning task in KG. The relations with less than 500 but more than 50 training triples are
        selected as testing data.
    \end{itemize}
    
    
    \subsection{Baselines}
    \label{baselines}
    
    We 
    adopt two embedding-based models, DistMult~\cite{yang2014embedding} and ConvE~\cite{Dettmers2018Convolutional}, as the base models 
    of our proposed modules,
    and compare the two enhanced models with the following popular relation prediction models:
    RESCAL~\cite{nickel2011three}, 
    TransE~\cite{bordes2013translating},
    DistMult~\cite{yang2014embedding}, 
    ComplEx~\cite{trouillon2017complex}, 
    ConvE~\cite{Dettmers2018Convolutional}, 
    ConvKB~\cite{nguyen-etal-2018-novel},
    D4-STE, D4-Bumbel~\cite{xu-li-2019-relation}, and
    TuckER~\cite{balazevic-etal-2019-tucker}.
    Besides the above 
    general models, we 
    test two additional models, GMatching~\cite{Xiongoneshot} and CogKR~\cite{du2019cogkr}, which are designed specifically for the few-shot relation learning.

    \subsection{Experimental Configuration}
    \label{settings}
    We implement the base models and our proposed two modules in PyTorch~\cite{paszke2017automatic}
    .
    Throughout the experiments, we 
    optimize the hyperparameters in a grid search setting for the best mean reciprocal rank~(MRR) 
    on the validation set.
    We use Adam to optimize all the parameters with initial learning rate at 0.003. 
    The dimensions of entity and relation embeddings 
    are both set to 200. 
    The loss weight~$\lambda$ 
    is set to 0.1.
    According to the frequency of relations,
    we 
    take the top 20\% and bottom 80\% of relations as many-shot and few-shot relation classes, respectively.
    The experimental results of our model 
    are averaged across three training repetitions, and standard deviations~(sd) are 
    also reported.  
    

    \subsection{Experiment \uppercase\expandafter{\romannumeral1}: Link Prediction}
    \label{exp1}
    
    \subsubsection{Setting}
    We follow the evaluation protocol of~\cite{Dettmers2018Convolutional}:
    each 
    input~$(e_{h},r,e_{t})$ is converted 
    to two queries, that is, tail query~$(e_{h},r,?)$ and head query~$(?,r,e_{t})$; then, 
    the ranks of correct entities are recorded 
    among all entities for each query, excluding 
    other correct entities 
    that were observed in 
    any of the train/valid/test sets for the same query.
    We use the 
    filtered HITS@1, 5, 10, and MRR as evaluation metrics. 

    \subsubsection{Results}
    Table~\ref{table:mainresult} 
    records the results on 
    five imbalanced datasets, which 
    reflect the general performance of the compared models in solving the link prediction task.
    It shows that 
    two base models~(DistMult and ConvE) 
    are generally 
    improved by 
    incorporating the proposed GRL framework. 
    That is, GRL 
    improves DistMult by an average of 3.84\% and improves ConvE by an average of 1.08\% under the MRR 
    evaluation.
    Especially, the enhanced model ConvE+GRL generally outperforms 
    the compared models 
    on YAGO3-10, FB15K-237, Kinship, and WN18, and the enhanced model DistMult+GRL 
    also performs well 
    on NELL-995.
    We also evaluate 
    the performance of GRL 
    in learning many-shot and few-shot relations and show the MRR results of DistMult, DistMult+GRL, ConvE, and ConvE+GRL on YAGO3-10 and NELL-995~(c.f. Table~\ref{table:headtail}).
    The results indicate that GRL achieves consistent improvements on both ``many-shot'' and ``few-shot'' sub-groups.
    We assume this may be because 
    handling many-shot relations can be improved thanks to useful implicit information from few-shot relations, even though 
    there are already
    numerous training samples for many-shot relations. 
    From this aspect, it is sensible for the mainstream link prediction models to rely on GRL regarding the imbalanced relation issue.

    

    
    

    \begin{table}[t]
        \begin{center}
        
            \small
            \resizebox{!}{!}{
                \begin{tabular}{lm{0.58cm}<{\centering}m{0.58cm}<{\centering}m{0.58cm}<{\centering}m{0.58cm}<{\centering}}
                
                \toprule[1.2pt]
                & \multirow{2}{*}{MRR} &  \multicolumn{3}{c}{HITS@N} \\ \cline{3-5}\rule{0pt}{10pt}
                &   &   @10& @5&@1\\
                \midrule[1.2pt]
                TransE${}^\dagger$	&	9.3	&	19.2	&	14.1	&	4.3	\\
                
                GMatching${}^\dagger$ 	&	18.8	&	30.5	&	24.3	&	13.3	\\
                CogKR${}^\ast$	&	\textbf{25.6}	&	35.3	&	31.4	&	\textbf{20.5}	\\
                \midrule[1.2pt]
                DistMult${}^\dagger$		&	10.2	&	17.7	&	12.6	&	6.6\\
                \multirow{1}{*}{DistMult+GRL}&	14.4	&	23.0	&	18.2	&	9.8\\
            ($\pm$ sd)&{ \small(2.0)}&{ \small(2.1)}&{ \small(1.9)}&{ \small(2.3)}\\
            \midrule[0.6pt]
                ConvE${}^\ast$ &	17.0	&	30.6	&	23.0	&	10.5\\
                \multirow{1}{*}{ConvE+GRL}&	\textbf{25.6}	&	\textbf{38.9}	&	\textbf{33.6}	&	{18.8}\\
                ($\pm$ sd)&{ \small(2.3)}&{ \small(3.7)}&{ \small(3.1)}&{ \small(2.1)}\\

                \bottomrule[1.2pt]
                
                \end{tabular}
                }
        \end{center}
        \caption{Few-shot relation learning results~(mean $\pm$ sd) on NELL-ONE dataset~(\%): the results marked by `$\dagger$' or `$\ast$' are taken from~\cite{Xiongoneshot,du2019cogkr}. 
        }
        \label{table:fewresult} 
    
        \end{table}

    \subsection{Experiment \uppercase\expandafter{\romannumeral2}: Few-shot Relation Learning}
    \label{exp2}
    
    \subsubsection{Setting}
    To further evaluate the 
    performance of GRL
    in the few-shot relation
    learning case, which is tricky for a link prediction model, especially, when relations are very insufficient, we 
    test approaches on the NELL-ONE dataset wherein each test relation has only one instance in the training set.
    We follow the evaluation protocol and metrics of~\cite{Xiongoneshot}.

    \subsubsection{Results}
    Table~\ref{table:fewresult} shows that GRL consistently improves both of the base models by average 4.2\% and 8.6\% MRR scores. Especially for ConvE, incorporating GRL 
    helps it outperform the other approaches on three metrics.
    CogKR, 
    a path learning based model, performs best under HITS@1. The reason might be that the testing query is easy to be completed by finding KG paths on the few-shot relation datasets, such as NELL-ONE. 
    Although there is only one training instance for each testing query, GRL can effectively embed the few-shot relations by learning from the semantically similar relations in the many-shot class.

    \subsection{Experiment \uppercase\expandafter{\romannumeral3}: Zero-shot Relation Learning}
    \label{exp3}
    
    \subsubsection{Setting}
    To evaluate the performance on zero-shot relations of GRL, we construct a testing set containing 500 triples whose relations are unseen in the training phase.
    The testing triples are randomly sampled from the FB15K dataset~\cite{bordes2013translating}, and the training set is FB15K-237 to ensure the authenticity of the triples.
    We adopt the fundamental testing protocol that quantitatively determines the scores of triples with zero-shot relations. 
        
        Most of existing zero-shot relation studies have to depend on textual descriptions, while the zero-shot learning addressed in this work does not require this information.
        Therefore, we select the GMatching model~\cite{Xiongoneshot} for comparison, which can predict similar relations by learning a matching metric without any additional information.
        We use the classical method TransE~\cite{bordes2013translating} to learn the relation embeddings in the FB15K dataset and calculate the similarity 
        between the zero-shot relation and the predicted relation.

    \begin{figure}[t]
        \centering
        \includegraphics [width=8.5cm]
        {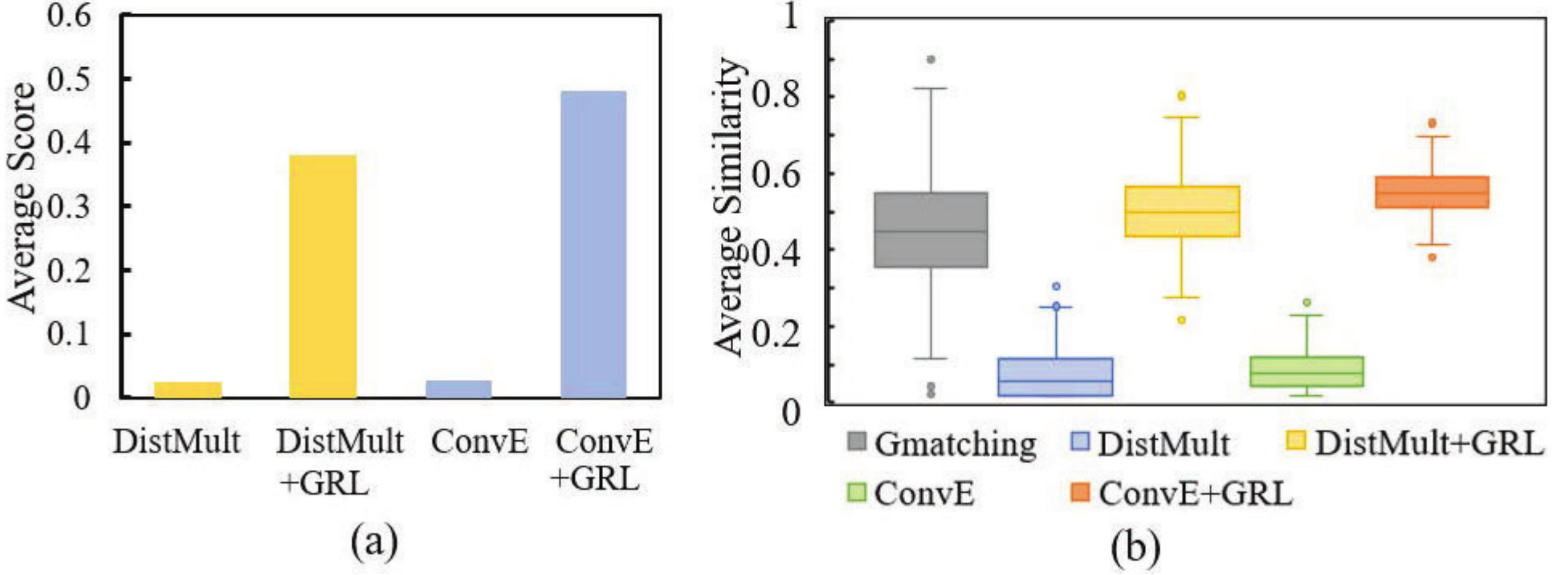}
        \caption{
        Zero-shot relation learning results: 
        (a) the average score of the testing triples,
        and (b) the average similarity between the zero-shot relation with the predicted relation.
        }
        
        \label{fig:unseen}      
        \end{figure}

    \subsubsection{Results}
        
        Figure~\ref{fig:unseen} (a) 
        demonstrates results of the average score of the testing triples with zero-shot relations.
        Note that we use the fusion vector as the zero-shot relations embedding.
        We can see that two base models (DistMult and ConvE) cannot get a good average score because all zero-shot relations will be identified as an `unknown' relation.
        When 
        GRL is plugged, two enhanced models (DistMult+GRL and ConvE+GRL) are both 
        boosted in learning positive relations, proving that the GRL framework can effectively improve the 
        validation capabilities on triples with zero-shot relations of the base models.
        Figure~\ref{fig:unseen} (b) shows the performance on predicting zero-shot relations. 
        We can see that the base models 
        perform worse due to their superficial way of embedding zero-shot relations as mentioned before.
        When equipping with GRL, the enhanced models perform better than GMatching, 
        indicating that learning from the semantic correlations between unseen relations and seen relations provides a comparably good way as learning from neighbor information.

    \begin{table}[t]
    
        \begin{center}

\small
    
        \begin{tabular}{m{0.38cm}<{\centering}lm{1.5cm}<{\centering}|m{1.58cm}<{\centering}}
            \toprule[1.2pt]
                          && YAGO3-10 & NELL-ONE \\  \midrule[1.2pt]
        (1)&ConvE             & 52.0     & 17.0     \\ \midrule[0.6pt]
        (2)&ConvE+GRL($p_{f}=0$)   & 52.6     & 23.3     \\ 
        (3)&ConvE+GRL($p_{f}=0.5$) & 53.9     & 24.7     \\ 
        (4)&ConvE+GRL($p_{f}=1$)   & 52.2     & 20.3     \\\midrule[0.6pt]
        (5)&ConvE+Direct      & 51.2     & 10.5     \\\midrule[0.6pt]
        (6)&ConvE+GRL         & 55.4     & 25.6    \\
        \bottomrule[1.2pt]
        \end{tabular}
        \caption{Ablation Study.}
        \label{table:abalationstudy} 
    \end{center}

    \end{table}

    \section{
    Further Analysis of GRL}
    \label{ablation}
    
    \subsection{Ablation Study} 
    \textbf{Study of Fusion Probability} 
    To assess the effect of the fusion vector, we 
    make a comparison on three variants from the fusion probability perspective based on ConvE, see Table~\ref{table:abalationstudy} (2)-(4).
    The three variants are the followings: 
    only using the joint vector~(i.e., $p_{f}=0$),
    only using the relational knowledge vector~(i.e., $p_{f}=1$), and 
    using the joint and relational knowledge vectors with an equal weight~(i.e., $p_{f}=0.5$).
    Compared with three variants, fusing the joint and relational knowledge vectors~(i.e., ConvE+GRL) performs best, which suggests the semantic correlations in the relational knowledge vectors can help the base model learn more appropriate representations of relations and thus boost the general performance. Moreover, the adaptive fusion probability can improve the flexibility of the fusion operator.


    \noindent
    \textbf{Direct Fusion vs. GRL}
    We test now a direct fusion method that 
    fuses the relational knowledge vector with the relation representation
    as the updated relation representation without considering the classification-aware loss.
    Table~\ref{table:abalationstudy} (5) shows the MRR performance of ConvE when enhanced by the direct method.
    Rich semantic correlations in KGs cannot be adequately learned by the direct method because it simply leverage the superficial semantic correlations, rather than embedding them into relation vectors.
    Moreover, the direct method will make embedding learning more confusing especially for the few-shot relation data such as NELL-ONE.
    
    \subsection{Case Study} 
    
\begin{figure}[t]
    \centering
    \includegraphics [width=8.5cm]
    {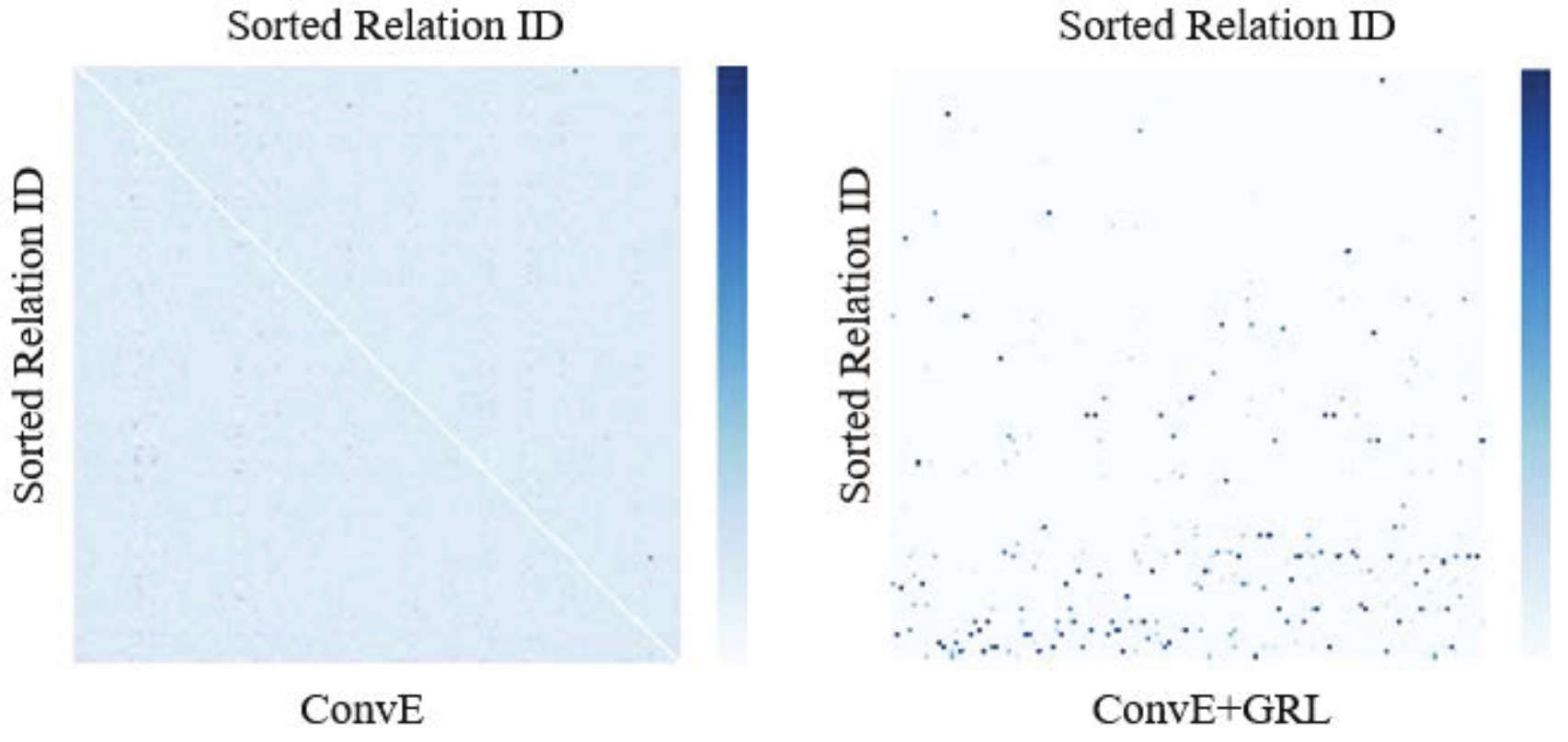}
    \caption{Case study: knowledge-aware attention cases with a heat map.}
    \label{fig:case1}
    
    \end{figure}

    \noindent
    \textbf{Visualization of Knowledge-aware Attention} 
    GRL 
    is able to make the base model fully learn semantic correlations between relations.
    To verify this, 
    we display the attention distribution 
    for the base model~(ConvE) and 
    enhanced model~(ConvE+GRL) on FB15K-237 in Figure~\ref{fig:case1}, and show the average attention distribution of 237 relation classes where each row represents a type of relation.
    The base model learns little about semantic correlations between relations,
    while the enhanced model~(ConvE+GRL) can perfectly capture the semantic correlations.
    The attention distribution of few-shot relations is more discrete than many-shot relations due to insufficient training data.
    
        \begin{figure}[t]

        \centering
        \includegraphics [width=8.5cm]
        {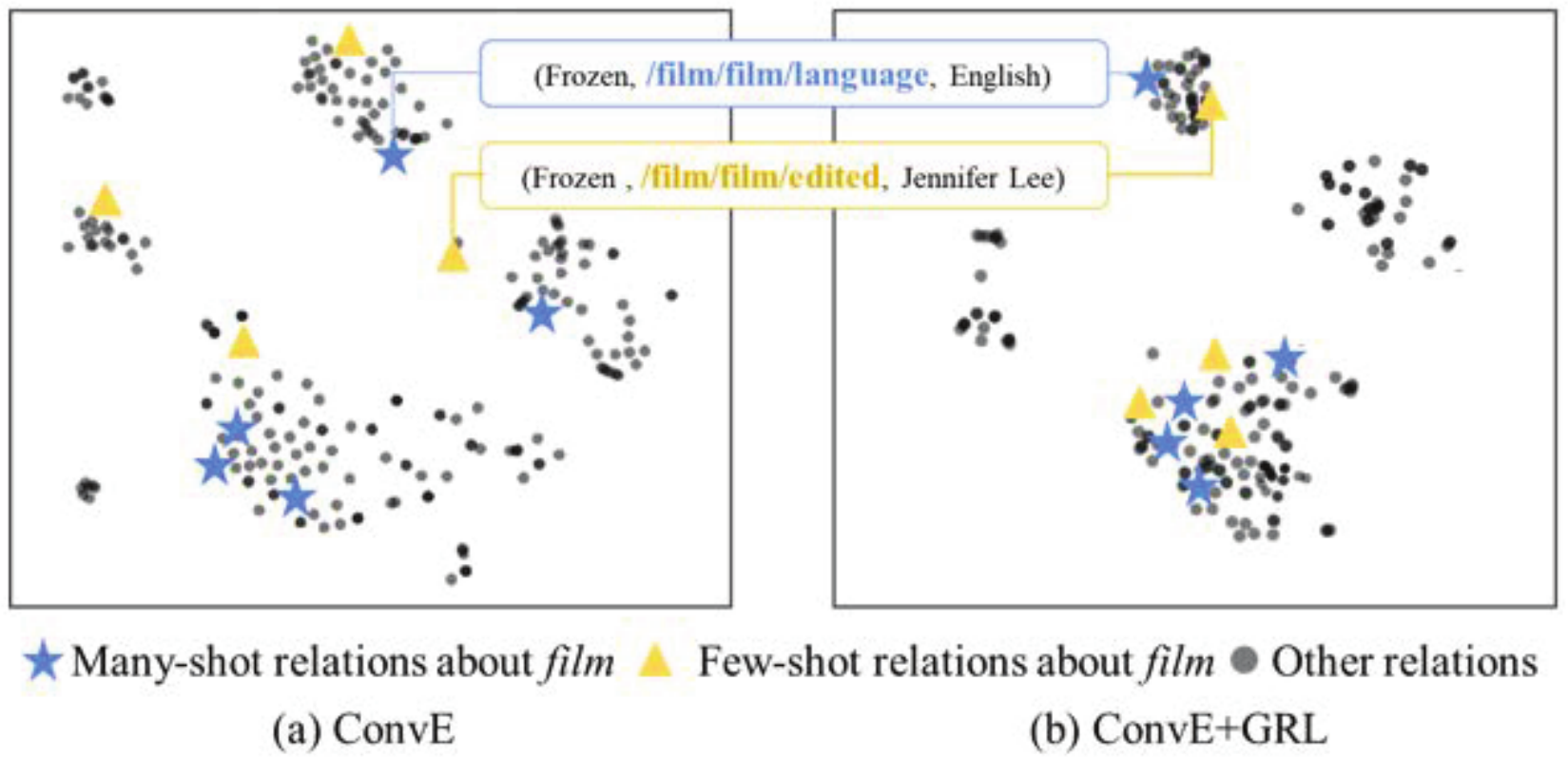}
        \caption{Case study: t-SNE visualization of relation embeddings in FB15K-237 (better view in color).
        The semantically similar relations get closer after plugging GRL.
        }
        \label{fig:intro2}      
        \end{figure}

    \noindent
    \textbf{Visualization of Relation Embedding} 
    In addition, we also show in Figure~\ref{fig:intro2} the t-SNE~\cite{maaten2008visualizing} plot of all relations on FB15K-237 in embedding space.
    To provide more insights we highlight the relations associated with ``film''.
    The Stars and Triangles represent the many-shot and few-shot relations, respectively.
    We can see that the many-shot and few-shot relations are more compact in the case of the enhanced model than the base model
    .

    \section{Conclusion and Future Work}
    \label{conclusion}
    In this work, we study two natural problems in the link prediction task:
    1) unbalanced relation distribution, and 2) unseen relations.
    To address them, we focus on generalized relation learning and propose a framework, GRL, that uses semantic correlations among relations as a bridge to connect semantically similar relations.
    Through extensive experiments on six datasets, we demonstrate the effectiveness of GRL, providing a comprehensive insight into the generalized relation learning of KGs. There are a few loose ends for further investigation. We will consider combining the external text information and the semantic knowledge of KGs to facilitate the relation learning.
    We will also try to deploy GRL to downstream applications that involve generalized relation learning scenarios to gain more insights.\\

    \section*{Acknowledgments} 
    This work was supported in part by the Ministry of Education of Humanities and Social Science project under grant 16YJC790123 and the Natural Science Foundation of Shandong Province under grant ZR2019MA049.

\bibliography{reference-short-nourl}

\end{document}